\documentclass[runningheads]{llncs}

\usepackage{eccv}

\usepackage{eccvabbrv}

\usepackage{graphicx}
\usepackage{booktabs}

\usepackage[accsupp]{axessibility}  % Improves PDF readability for those with disabilities.

\usepackage[breaklinks,colorlinks,citecolor=eccvblue]{hyperref}
\usepackage{orcidlink}

\usepackage{tikz,bm}

\usepackage[table]{xcolor}
\definecolor{oursgray}{gray}{0.93}
\usepackage{algorithm}
\usepackage{algpseudocode}
\usepackage{amsmath,amssymb,bm,multirow}
\usepackage{booktabs,makecell,array,caption}
\usepackage{wrapfig}

\begin{document}

\title{SR-Edit: Region-Aware Image Editing via Self-Refinement} 

% \titlerunning{SR-Edit: Region-Aware Image Editing via Self-Refinement}

\author{
Andong Wang\inst{1,2}\orcidlink{0009-0008-0567-4940}
\and
Zehua Chen\inst{1}\orcidlink{0009-0000-2294-7447}\thanks{Corresponding authors: Zehua Chen and Jun Zhu.}
\and
Yuxuan Jiang\inst{1}\orcidlink{0009-0003-7134-7810}
\and
Jun Zhu\inst{1}\orcidlink{0000-0002-6254-2388}\ensuremath{^\star}
}

\authorrunning{Wang et al.}

\institute{Tsinghua University, Beijing, China
 \and Renmin University of China, Beijing, China
 \\\email{andong000@ruc.edu.cn \ \{zhc23thuml,dcszj\}@mail.tsinghua.edu.cn}
}

\maketitle

\begin{abstract}
With the recent rapid progress in generative models, image editing has made remarkable advances, yet achieving faithful edits that precisely modify only the target regions while strictly preserving all other regions remains challenging. 
Since externally provided region annotations are often difficult to obtain in practice, a growing body of work seeks to improve preservation by automatically inferring edit and non-edit regions, and then enforcing consistency on the latter. 
However, these approaches still suffer from inaccurate region estimation and heuristic correction strategies that distort the native inference process, making methods designed for fidelity themselves a new source of artifacts. 
We propose SR-Edit, an image editing framework that overcomes these issues via iterative self-refinement. Specifically, at each iteration, SR-Edit first (i) extracts progressively precise and self-consistent region separation from the model’s own predictions by lightweight post-processing, and then (ii) enforces preservation in non-edit areas through correction updates that remain aligned with the original sampling dynamics.
Extensive experiments demonstrate that SR-Edit achieves superior preservation and overall image quality compared to existing editing techniques.
\keywords{Image Editing \and Generative Models \and \textit{h}-transform theory}
\end{abstract}
\begin{figure}[h!]
    \centering
    \includegraphics[width=0.91\linewidth]{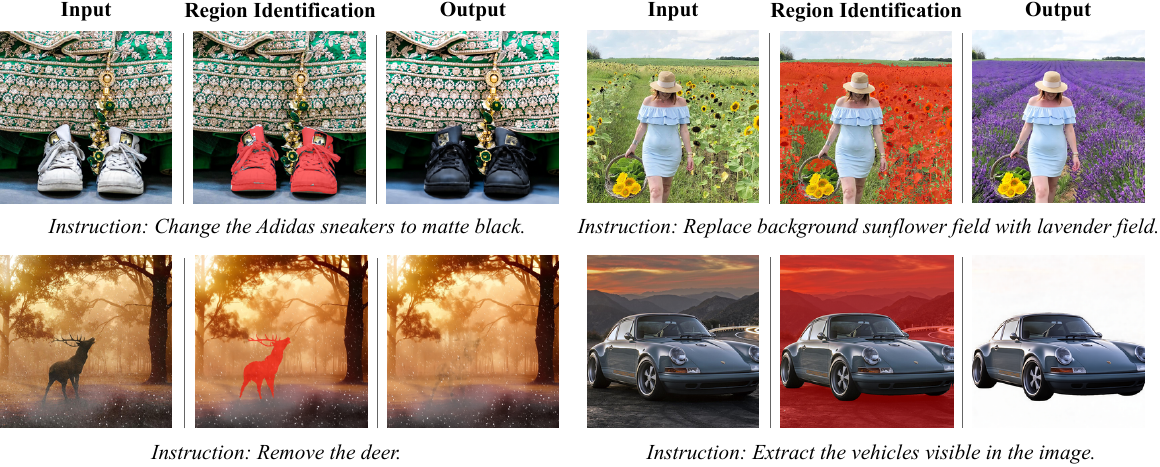}
    \caption{\textbf{Examples of edited images produced by SR-Edit}. The edit regions are automatically inferred by the model and indicated by red overlays.}
    \label{fig:demo}
\end{figure}

\section{Introduction}
\label{sec:intro}

Recent progress in generative models~\cite{ddpm,sde,flow-matching,reci-flow,img-ldm} has greatly expanded the capabilities of image editing. A variety of editing paradigms have been proposed, including editing with forward noising followed by conditional denoising~\cite{sdedit}, training-based editors that learn an explicit mapping from a source image and a textual editing instruction to the edited image~\cite{pix2pix,omniedit,anyedit,step1x,qwen-image}, and inversion-based methods that invert an image into a diffusion trajectory and then modify the denoising process to achieve the edit~\cite{p2p,null-text,imagic,kv-edit,eedit,follow-your-shape}.

Despite impressive results, a persistent bottleneck remains: \emph{faithful editing}, i.e., producing the desired modifications while \emph{strictly preserving} all other content.
In real-world edits, users typically expect backgrounds, identity cues, fine textures, and geometric details outside the target concept to remain unchanged.
However, most generative editing pipelines still rely on sampling dynamics that start from a noisy state and evolve under a new condition, which inherently operates on global image content~\cite{sdedit}.
As a result, even when the semantic edit succeeds, the output may exhibit unintended drift in non-edited regions, such as subtle background changes, texture inconsistencies, or identity degradation.

A common strategy to improve preservation is to explicitly separate the image into \emph{edit} and \emph{non-edit} regions, and then enforce consistency on the latter.
When accurate user-provided masks are available, spatially constrained editing can be effective~\cite{repaint,blend,emu-edit}, but manual annotation is costly and often unavailable in practice.
This motivates mask-free approaches that infer editable regions automatically, for example by manipulating or interpreting attention maps~\cite{p2p,pix2pix-zero,pnp,lime,concept-attn} or by computing differences between generation trajectories under source and target prompts~\cite{diff-edit,follow-your-shape}.

Nevertheless, automatic region estimation remains challenging. The inferred masks are often insufficiently precise, failing to align with true pixel boundaries, and can also be temporally unstable across sampling steps~\cite{masactrl}.
Moreover, many preservation mechanisms rely on heuristic interventions such as feature fusion and reuse, or KV~\cite{attention} interpolation in intermediate representations, which are not consistent with the original sampling dynamics~\cite{spot-edit,follow-your-shape}.
Consequently, the techniques introduced above to improve fidelity do not reliably enhance preservation. Instead, they can become a new source of imprecision and artifacts.

\begin{figure}[t!]
    \centering
    \includegraphics[width=0.93\linewidth]{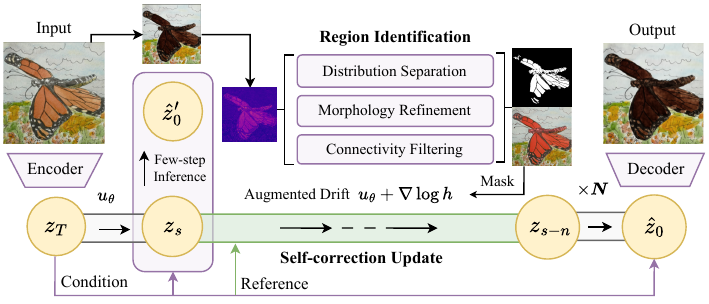}
    \caption{\textbf{Overview of SR-Edit}. The edit instruction of the case is \emph{Change the animal's fur color to a darker shade}. We use $u_\theta$ to denote the sampling vector field that drives the generative trajectory. $s$ denotes the start time of self-correction update while $n$ denotes the time duration. $N$ denotes the number of self-refinement iterations.}
    \label{fig:over}
\end{figure}

We propose \textbf{SR-Edit}, a \emph{self-refined} image editing framework that addresses both failure modes: inaccurate region estimation and heuristic preservation corrections.
The key idea is to \emph{use the model's own output as self-feedback} to construct a reliable pixel-space separation between edited and preserved content, and then to enforce preservation through an update rule that remains aligned with intrinsic sampling dynamics.

Concretely, SR-Edit leverages the model's predictions to build \emph{self-consistent difference maps} and converts them into a precise edit and non-edit decomposition via lightweight pixel-space post-processing~\cite{morphological,morphological2}.
Unlike latent-space discrimination, pixel-space differences directly reflect decoded visual changes and admit mature refinement operations, yielding sharper and more stable region estimates.

To enforce preservation, we adopt a principled conditioning mechanism based on \emph{Doob's $h$-transform} theory~\cite{doob,sb}.
We formulate preservation as a constraint on the non-edit region by modeling it as a vanishing-noise observation, and derive an additive guidance term that enhances fidelity while preserving the structure of the sampling procedure.
Under a plug-in approximation, this guidance reduces to a masked residual on the model's clean-image prediction, which can be injected into diffusion samplers as an additive drift correction and analogously into flow-matching samplers as a velocity correction.

In contrast to heuristic state interventions (e.g., direct feature replacement or blending)~\cite{spot-edit,attention}, our update is derived from a distribution transform and therefore tends to preserve native inference dynamics while suppressing non-edit drift.
This complements existing methods that improve fidelity to the source image through more accurate inversion~\cite{null-text,edict,structure-preserve} and structural conditioning approaches that anchor geometry of the source via external controls~\cite{controlnet,omniedit}.

SR-Edit operates in an \emph{iterative self-correction} loop: region separation is inferred from current predictions, preservation guidance is applied to suppress non-edit drift, and the model produces a progressively refined result across iterations.
This self-refinement gradually improves region localization and preservation while maintaining edit intent.
Extensive experiments demonstrate that SR-Edit yields superior non-edit preservation and improved overall visual quality compared to prior region-aware and mask-free editing techniques~\cite{diff-edit,follow-your-shape,spot-edit}.

Our contributions can be summarized as follows:
\begin{itemize}
    \item We introduce \textbf{SR-Edit}, an image editing framework that improves faithfulness by jointly enhancing region identification and preservation in an iterative self-refinement process.
    \item We propose \textbf{self-consistent region identification} derived from the model's own predictions and show that \textbf{pixel-space} post-processing enables more precise and stable edit and non-edit separation than latent or attention-based discrimination.
    \item We develop a \textbf{Doob's $h$-transform} formulation for region preservation and derive a practical masked guidance rule that integrates naturally with both diffusion and flow-matching samplers.
    \item We validate that SR-Edit improves non-edit preservation and overall image quality across diverse editing scenarios, outperforming existing editing techniques.
\end{itemize}

\section{Related Work}
\label{sec:related_works}

\subsection{Image Editing with Generative Model Backbones}

\paragraph{Generative model backbones.}
Modern generative models, such as diffusion models~\cite{ddpm,sde,img-ldm}, flow matching~\cite{flow-matching,reci-flow,flux1kontext}, and bridge models~\cite{ddbm,bridge-tts,AudioLBM,GuidedBridge}, have shown a strong capability of faithfully reconstructing a target distribution with learned time-dependent scores or vector fields.
Given a scalable network architecture~\cite{img-ldm,SiT,AudioLDM}, these generative frameworks have been able to capture complex data distributions defined by large-scale datasets, enabling high-fidelity generation across data modalities, such as image~\cite{img-ldm,StableDiffusion3}, audio~\cite{InferGrad,BinauralGrad,FreeAudio}, video~\cite{FrameBridge,DiffGAP,VideoDiffusion}, or time-series signals~\cite{RespDiff,RefineBridge,chen2026versatile}.
In the inference process, these frameworks usually start from a prior distribution,~\textit{e.g.}, Gaussian noise or a clean representation, and gradually generate the target with iterative refinement steps, showing a noise-to-data~\cite{Omni2Sound,ControlAudio,AudioMoG} or data-to-data sampling trajectory~\cite{Bridge-SR,VoiceBridge}. 

\paragraph{Image editing paradigm.} Following the remarkable breakthroughs of generative models, a growing body of research has focused on adapting these models for editing tasks~\cite{survey,FreeSonic}. 
Image editing methods can be roughly grouped by how they incorporate an input image and enforce edit intent. 
\emph{Forward-and-backward} approaches inject noise into the input and denoise under new conditions, using the noise level to regulate the edit magnitude~\cite{sdedit}.
\emph{Training-based instruction editing} methods learn an explicit mapping from the source image and textual instruction to the edited output through supervised or synthetic training pipelines, allowing direct and scalable manipulation~\cite{pix2pix,omniedit,anyedit,step1x,qwen-image}. \emph{Inversion-based} methods first invert a real image into a diffusion latent trajectory and then apply prompt changes while trying to preserve reconstruction~\cite{p2p,null-text,imagic,kv-edit,eedit,follow-your-shape}.

\paragraph{Preservation of the source.} Preserving the source image during edits is still difficult because noisy initialization in generative backbones can induce global drift, causing unintended changes in irrelevant regions. Existing approaches improve preservation through several typical mechanisms. \cite{null-text,edict,structure-preserve} achieve {high-fidelity inversion and reconstruction}, then the conditional sampling can start from a more faithful state and thus reduce global deviation. 
External or automatically estimated {masks separate edit regions and non-edit regions}, which explicitly improve preservation of the source~\cite{blend,diff-edit}. We will further discuss the method in Sec.~\ref{related_work:region}. 
Manipulating attention features~\cite{attention} or injecting intermediate features can also enforce spatial alignment between the source image and edited output~\cite{p2p,pix2pix-zero,pnp,masactrl}. Moreover, {external structural constraints} can be conditioned on edges, poses, segmentation, or depth, providing an explicit control to preserve original structure even when appearance changes \cite{omniedit,controlnet}.

\subsection{Region-Aware Image Editing}
\label{related_work:region}

\paragraph{External mask.} Accurate edit and non-edit region separation offers an important pathway to faithful image editing. Early and widely-adopted approaches rely on binary masks provided by users to explicitly specify where edits should occur~\cite{repaint,blend,emu-edit}.
This design provides direct spatial control and often yields strong results when accurate masks are available.
However, it requires manual annotation, which limits applicability in real-world usage.
Moreover, the manual mask often has an inaccurate boundary, leading to incorrect modifications on irrelevant regions and therefore degrading edit faithfulness.

\begin{figure}[t!]
    \centering
    \includegraphics[width=1\linewidth]{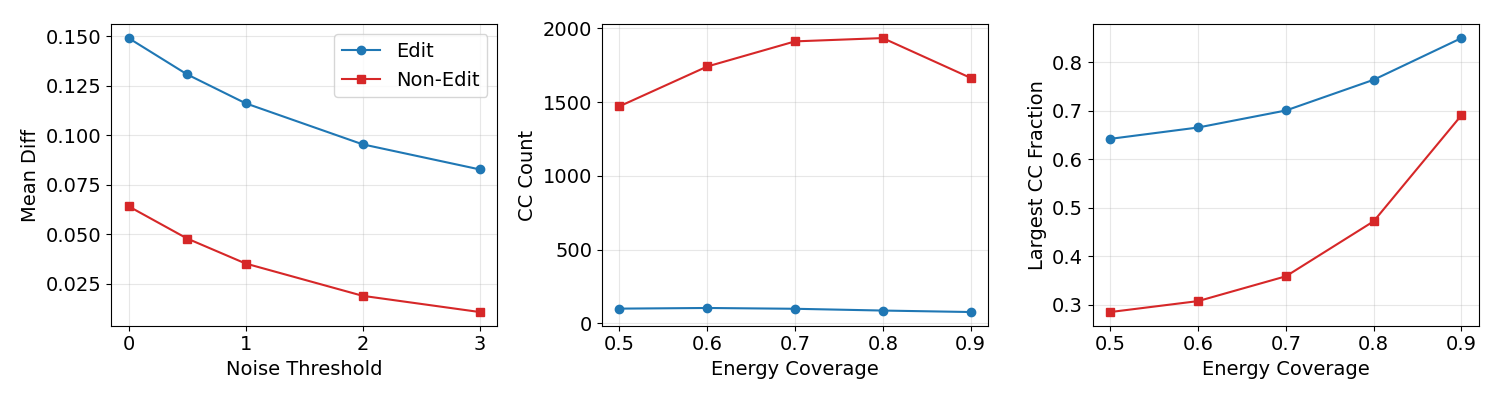}
\caption{
\textbf{Quantitative evidence of model behavior.} 
On edit cases and human-annotated region partitions from MagicBrush test split~\cite{magicbrush}, editing model~\cite{qwen-image} shows a clear pattern: edit regions concentrate strong changes in a coherent area, while non-edit regions produce low-magnitude and weakly structured changes.
\textbf{Left:} mean pixel change inside and outside the annotated region after applying a MAD-based threshold\protect\footnotemark[1]; the inside-to-outside ratio increases from $2.32$ to $7.71$, indicating much higher change energy in edit regions.
\textbf{Center:} number of connected components among pixels covering a top fraction of total change energy, extremely large counts (e.g., $1472$ vs.\ $100$ at $0.5$ coverage) indicate that non-edit differences are highly fragmented whereas edits remain spatially concentrated.
\textbf{Right:} fraction of selected pixels contained in the largest connected component, further confirming stronger spatial concentration for edits (e.g., $64\%$ vs.\ $29\%$ at $0.5$ coverage).
}

\label{fig:change_concentration}
\end{figure}
\footnotetext[1]{MAD~\cite{mad,alter-mad} (median absolute deviation), a robust noise scale estimator that reduces the impact of outliers by using the median and absolute deviations.}

\paragraph{Text-driven localization.} To reduce reliance on external masks, subsequent work investigates inferring editable regions from text-conditioned signals alone.
DiffEdit and Follow-Your-Shape~\cite{diff-edit,follow-your-shape} derive edit regions by computing differences between inference processes conditioned on the source and target prompts.
Image editing frameworks based on manipulation of cross-attention maps associate textual tokens with spatial regions, enabling localized edits without external masks~\cite{pix2pix-zero,lime,p2p,concept-attn}.
However, the editable region is inferred from semantic representations instead of pixel-level evidence, and the subsequent projection from semantic differences to image region separation is only approximate, often resulting in imprecise and unstable localization~\cite{masactrl}.

\paragraph{Region identification via inference behavior.} More recent methods aim to localize editable regions by directly leveraging the model's inference behavior, which is naturally aligned with the editing process.
SpotEdit~\cite{spot-edit} further explores region control from the inference trajectory by leveraging single-step reconstruction and performing linear interpolation between KV features~\cite{attention} of intermediate latents and references.
Although effective in practice, the single-step reconstruction is typically sensitive, while the reliance on KV-level masking or interpolation introduces heuristic engineering choices that could be refined toward more principled and precise localization.

\section{SR-Edit}

\begin{figure}[t!]
    \centering
    \includegraphics[width=\linewidth]{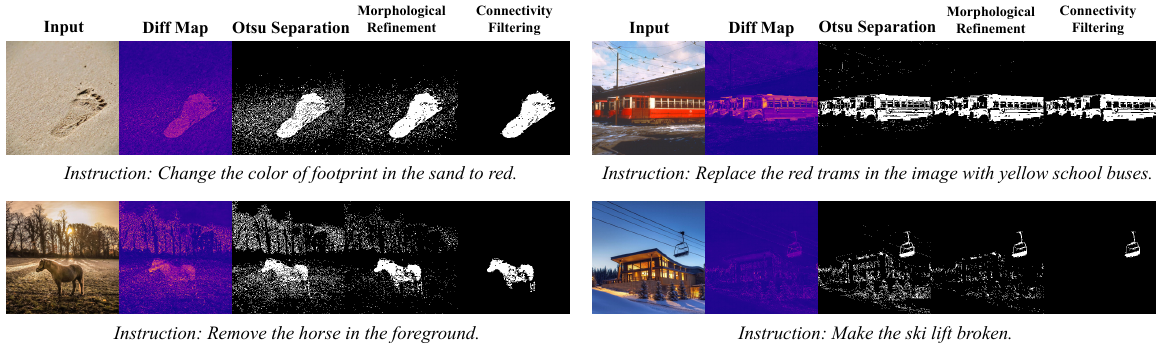}
    \caption{\textbf{Visualization of the incremental effects of each component in the SR-Edit post-processing pipeline for region identification.}}
    \label{fig:post-process-effect}
\end{figure}

At inference time, we organize editing into a short \emph{stabilization} stage followed by a stack of \emph{SR-Edit blocks}.
Each SR-Edit block performs a lightweight \emph{probe-and-refine} cycle: (i) a few-step inference produces a provisional edit $\mathbf{I}_{\mathrm{ref}}$, from which we infer the edit and non-edit regions via the pixel-space pipeline in Sec.~\ref{sec:region-id}; (ii) conditioned on this newly estimated region, we run a few guidance steps using the Doob's $h$-transform term in Sec.~\ref{sec:doob_diffusion} to enforce region preservation while continuing the edit.
Stacking these blocks yields an iterative self-refinement process: the model repeatedly \emph{re-estimates} where changes occur and \emph{re-applies} principled preservation guidance, so the region identification is continually refined along with the prediction, progressively sharpening region separation for more accurate boundaries while keeping the edit flexible.

\subsection{Precise Region Identification}
\label{sec:region-id}

\paragraph{Observation.}
Modern training-based image editors~\cite{anyedit,step1x,qwen-image,flux1kontext} often demonstrate reasonably good fidelity in practice: meaningful modifications tend to concentrate in the intended edit region, while non-edit areas are largely preserved, with residual deviations typically small and weakly structured (Fig.~\ref{fig:change_concentration}). This empirical behavior makes the editor output itself a useful self-feedback cue for region separation: large and spatially coherent differences indicate edited areas, whereas scattered low-amplitude differences are more consistent with preserved regions.

Moreover, while many editing models operate in the latent space, performing region discrimination in decoded pixel space rather than latent space enables more precise region identification and allows us to leverage mature image post-processing techniques~\cite{morphological,morphological2} to suppress scattered artifacts, since latent features are typically patch-level and their variations are not a stable proxy for pixel changes due to highly nonlinear decoding~\cite{img-ldm}. 
Therefore, following prior practices~\cite{spot-edit,follow-your-shape,diff-edit} that exploit output-driven cues for refinement and motivated by the advantages of pixel-space analysis discussed above, we explicitly treat the model's own prediction as a stable self-feedback signal in pixel space to infer edit regions.

\paragraph{Region identification pipeline.}
Given an input image $\mathbf{I}_{\mathrm{in}}$ and an edited image $\mathbf{I}_{\mathrm{ref}}$ produced by few-step inference, both defined on the same pixel grid $\Omega=\{1,\dots,H\}\times\{1,\dots,W\}$ with $C$ channels, we estimate a binary change mask $\mathbf{M}:\Omega\rightarrow\{0,1\}$, where $\mathbf{M}(\mathbf{p})=1$ indicates edited pixels and $\mathbf{M}(\mathbf{p})=0$ indicates non-edited pixels. The method involves four steps: (i) constructing a difference map, (ii) binarization using Otsu's threshold, (iii) morphological refinement, and (iv) connected-component filtering. The incremental effects of each component can be viewed in Fig.~\ref{fig:post-process-effect}.

\paragraph{Difference map.}
We compute a difference map $d(p)$ by taking the per-pixel absolute difference between the input and reference images, averaging over the $C$ channels:
\begin{equation}
d(p) = \frac{1}{C} \sum_{c=1}^{C} \left|\mathbf{I}_{\mathrm{in}}(p,c) - \mathbf{I}_{\mathrm{ref}}(p,c)\right|, \qquad p \in \Omega.
\end{equation}
This map $d(p)$ quantifies the local change at each pixel, with larger values indicating greater deviation between the images.

\begin{figure}[t!]
    \centering
    \includegraphics[width=0.97\linewidth]{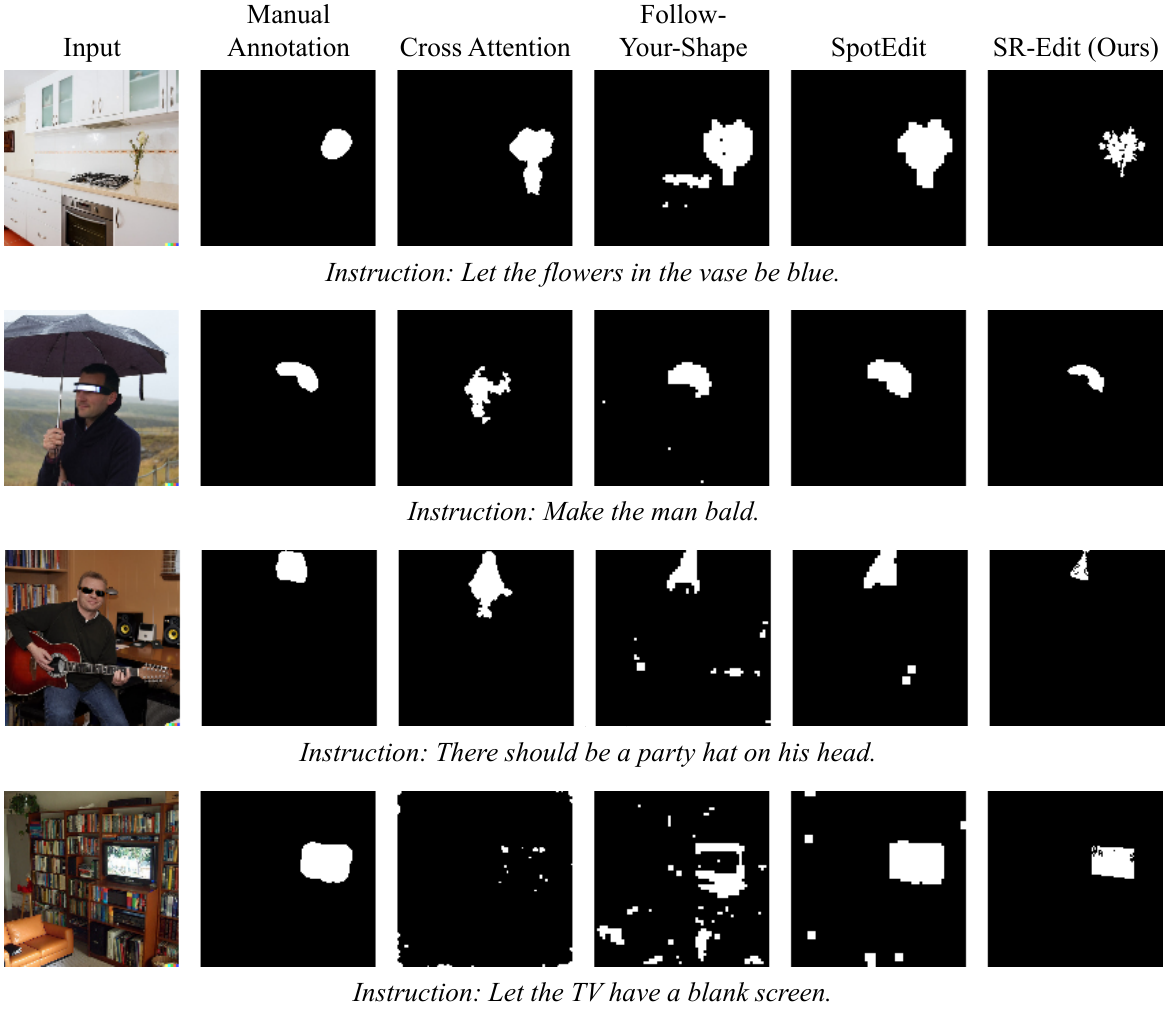}
    \caption{\textbf{Comparison on region identification between different methods.} We include region separation results from manual annotations in MagicBrush~\cite{magicbrush}, cross-attention maps~\cite{attention} on token with largest L2 energy, Follow-Your-Shape~\cite{follow-your-shape}, SpotEdit~\cite{spot-edit} and SR-Edit. All evaluated methods use Qwen-Image-Edit 2511~\cite{qwen-image} as the backbone.}
    \label{fig:mask-compare}
\end{figure}

\paragraph{Otsu thresholding.}
To binarize $d$ without manual tuning, we apply Otsu's separation method~\cite{otsu}. Its key idea is to choose a
single global threshold $t$ that best separates the histogram of $d$ into two classes (low differences as
\emph{unchanged} and high differences as \emph{changed}). Specifically, for each candidate threshold $t$, Otsu defines class
probabilities $\omega_0(t),\omega_1(t)$ and class means $\mu_0(t),\mu_1(t)$ from the histogram, and selects the
threshold that maximizes the between-class variance:
\begin{equation}
t^\star=\arg\max_t \ \omega_0(t)\,\omega_1(t)\big(\mu_0(t)-\mu_1(t)\big)^2.
\label{eq:otsu}
\end{equation}
We then obtain the initial mask by thresholding:
$\mathbf{M}_0(p)=\mathbb{I}\!\left[d(p)\ge t^\star\right]$, where $\mathbb{I}[\cdot]$ denotes the indicator function that returns $1$ if the condition holds and $0$ otherwise.

\paragraph{Morphological refinement.}
The initial mask $\mathbf{M}_0$ may contain small isolated artifacts and small holes or breaks inside true change regions.
We refine it using binary morphology~\cite{morphological,morphological2}.
Specifically, we use a small circular kernel on the pixel grid:
we first apply \emph{opening} with kernel size $s_{\mathrm{open}}$ to remove small foreground speckles, and then apply \emph{closing} with size $s_{\mathrm{close}}$ to fill small holes and connect narrow gaps.
This morphological refinement yields a cleaner and more spatially coherent mask $\mathbf{M}_1$.

\paragraph{Connected-component filtering.}
Finally, we run connected-component labeling on $\mathbf{M}_1$~\cite{digital-picture-processing}. Each connected component corresponds to a maximal set of foreground pixels that are
mutually reachable through neighbor-to-neighbor steps. We remove components whose area (pixel count) falls below a minimum threshold $A_{\min}$, treating them as residual noise, and keep the remaining components as the final change
mask $\mathbf{M}$. Our method shows improved precision in region identification compared with prior approaches, as illustrated in Fig.~\ref{fig:mask-compare}.

\subsection{Region Preservation via Doob's $h$-transform}
\label{sec:doob_diffusion}

Our method achieves region preservation by conditioning the sampling dynamics through an additive Doob's $h$-transform~\cite{doob,sb} guidance term.
In contrast to heuristic feature fusion such as KV reuse~\cite{attention}, the correction is derived from a distribution transform and thus more consistent with the underlying generative process.

\paragraph{Diffusion model.}
We consider a continuous-time diffusion model~\cite{ddpm} defined by the forward SDE~\cite{sde}
\begin{equation}
\label{eq:diff_forward_sde}
d\bm x_t=f(\bm x_t,t)\,dt+g(t)\,d\bm w_t.
\end{equation}
Here $\bm w_t$ is standard Brownian motion, $f$ is the drift, and $g(t)$ controls the noise scale.
Let $p_t$ denote the marginal density of $\bm x_t$ at time $t$.
For conditional editing, we condition on a control signal $\mathcal C$ and write the corresponding marginal as $p_t(\bm x\mid\mathcal C)$.
Its score $\nabla_{\bm x}\log p_t(\bm x\mid\mathcal C)$ is the conditional log-density gradient that drives the reverse-time denoising dynamics.

\paragraph{Observation model and Doob function.}
Let $\bar{\mathbf M}$ denote the complement of the mask defined earlier, and let $\bm y \triangleq \bar{\mathbf M}\bm x_{\mathrm{src}}$.
We enforce the preservation constraint in the form
$\bar{\mathbf M}{\bm x}_0=\bm y$.
We encode preservation with the vanishing-noise observation model $\bm Y=\bar{\mathbf M}\bm x_0+\bm\xi$, $\bm\xi\sim\mathcal N(\bm 0,\sigma_{\mathrm{obs}}^2I)$, and take $\sigma_{\mathrm{obs}}\to 0$.
The corresponding likelihood is $p(\bm y\mid \bm x_0)=\mathcal N(\bm y;\bar{\mathbf M}\bm x_0,\sigma_{\mathrm{obs}}^2I)$.
Define the Doob function
\begin{equation}
\label{eq:doob_h_def_diff}
h(t,\bm x)\triangleq \mathbb E\!\left[p(\bm y\mid \bm x_0)\,\middle|\,\bm x_t=\bm x,\mathcal C\right].
\end{equation}
By the Doob's $h$-transform, conditioning on $\bm Y=\bm y$ yields the factorization
$p_t(\bm x\mid \bm y,\mathcal C)\propto p_t(\bm x\mid \mathcal C)\,h(t,\bm x)$, and therefore the conditional score decomposes additively as
\begin{equation}
\label{eq:score_decomp}
\nabla_{\bm x}\log p_t(\bm x\mid \bm y,\mathcal C)
=
\nabla_{\bm x}\log p_t(\bm x\mid \mathcal C)
+
\nabla_{\bm x}\log h(t,\bm x).
\end{equation}
Consequently, any reverse-time diffusion sampler that uses the reference score can be made region-preserving by augmenting its score estimation with the guidance term $\nabla_{\bm x}\log h(t,\bm x)$.

\paragraph{Plug-in and masked Gaussian guidance.}
Exact evaluation of $h(t,\bm x)$ is generally intractable.
Let $\widehat{\bm x}_0(\bm x_t)$ be the model prediction of the clean image at time $t$.
We adopt a plug-in approximation that replaces the latent $\bm x_0$ inside the likelihood by $\widehat{\bm x}_0(\bm x_t)$, which gives
$h(t,\bm x_t)\approx \mathcal N(\bm y;\bar{\mathbf M}\widehat{\bm x}_0(\bm x_t),\sigma_{\mathrm{obs}}^2 I)$.
Equivalently, up to an additive constant independent of $\bm x_t$,
\begin{equation}
\label{eq:logh_plugin}
\log h(t,\bm x_t)
\approx
-\frac{1}{2\sigma_{\mathrm{obs}}^2}\,
\big\|\bar{\mathbf M}\big(\widehat{\bm x}_0(\bm x_t)-\bm x_{\mathrm{src}}\big)\big\|_2^2 .
\end{equation}
Differentiating yields
\begin{equation}
\label{eq:grad_logh_exact}
\nabla_{\bm x_t}\log h(t,\bm x_t)
\approx
-\frac{1}{\sigma_{\mathrm{obs}}^2}\,
J_{\widehat{\bm x}_0}(\bm x_t)^\top\,
\bar{\mathbf M}\big(\widehat{\bm x}_0(\bm x_t)-\bm x_{\mathrm{src}}\big),
\end{equation}
where $J_{\widehat{\bm x}_0}(\bm x_t)$ denotes the Jacobian of $\widehat{\bm x}_0$ with respect to $\bm x_t$.
In implementation, the local sensitivity is absorbed into a time-dependent scalar schedule, which yields the practical masked-residual form
\begin{equation}
\label{eq:grad_logh_practical}
\nabla_{\bm x_t}\log h(t,\bm x_t)
\approx
-\lambda(t)\,\bar{\mathbf M}\big(\widehat{\bm x}_0(\bm x_t)-\bm x_{\mathrm{src}}\big).
\end{equation}

Combining Equations~\eqref{eq:score_decomp} and \eqref{eq:grad_logh_practical} shows that, under the plug-in approximation, the $h$-transform guidance reduces to a masked residual on the model prediction $\widehat{\bm x}_0(\bm x_t)$.
This residual can be viewed as a time-consistent error signal, so the correction acts as a conditional energy tilt toward preservation rather than a heuristic state intervention~\cite{doob,sb,survey-sb}.
Since it enters only through the drift, it does not modify the diffusion coefficient $g(t)$ and therefore typically leaves the noise schedule and sampling structure largely intact.
In the ideal case where the base conditional dynamics already match the desired conditioned process, the guidance term vanishes.

\subsubsection{Preservation guidance for flow matching}
\label{sec:doob_flow}

Flow models~\cite{flow-matching,reci-flow} sample by the ODE $\frac{d\bm x_t}{dt}=\bm u_\theta(\bm x_t,t)$ and do not come with a canonical Doob's $h$-transform tied to an SDE.
After the same plug-in reduction, preservation is encoded by the quadratic energy~\cite{flow-guidance}
$E_t(\bm x_t)\triangleq \frac{1}{2\sigma_{\mathrm{obs}}^2}\|\bar{\mathbf M}(\widehat{\bm x}_0(\bm x_t)-\bm x_{\mathrm{src}})\|_2^2$,
whose gradient yields the same masked residual direction.
Because this guidance depends only on $\widehat{\bm x}_0(\bm x_t)$ and not on a diffusion coefficient, it can be injected as an additive velocity correction without changing the ODE form:
\begin{equation}
\label{eq:fm_guided}
\frac{d\bm x_t}{dt}
=
\bm u_\theta(\bm x_t,t)
-
\gamma(t)\,\nabla_{\bm x_t}E_t(\bm x_t)
\;\;\approx\;\;
\bm u_\theta(\bm x_t,t)
-
\gamma(t)\,\lambda(t)\,\bar{\mathbf M}\big(\widehat{\bm x}_0(\bm x_t)-\bm x_{\mathrm{src}}\big),
\end{equation}
which retains the flow matching transport structure while encouraging $\bar{\mathbf M}\widehat{\bm x}_0=\bm y$.

\section{Experiments}
\subsection{Dataset}
\label{sec:dataset}
We evaluate on \textbf{ImgEdit-Bench}, a benchmark derived from the ImgEdit dataset for instruction-based image editing~\cite{imgedit}. 
Since our method focuses on spatially localized editing, we focus on the \textbf{single-turn} setting and consider editing tasks confined to a specific area of the image, including \emph{replace}, \emph{adjust}, \emph{background}, \emph{remove}, \emph{add}, and \emph{action}, resulting in 497 valid cases in total.

We additionally evaluate on a widely used image-editing benchmark \textbf{PIE-Bench}~\cite{pie}, which provides human-annotated edit masks together with descriptive captions. The masks enable separate evaluation of edited and non-edited regions, while the captions allow semantic consistency between the edited image and the intended content. Together, these annotations provide complementary and more fine-grained evaluation dimensions beyond ImgEdit-Bench. As above, we exclude style edits and viewpoint transformations that cannot be meaningfully localized, resulting in 578 cases for evaluation.

\begin{table}[t]
\centering
\renewcommand{\arraystretch}{1.1}
\setlength{\abovecaptionskip}{8pt}
\setlength{\tabcolsep}{2pt}

\begin{tabular}{
  @{\hspace*{-0.5pt}}l@{\hspace*{-0.5pt}}
  *{4}{w{c}{1.03cm}}
  @{}p{12pt}@{}c
  @{}p{14pt}@{}c
}
\cmidrule[\heavyrulewidth](l{-2pt}r{-2pt}){1-9}

\multirow{2}{*}{\raisebox{-0.6ex}{\textbf{Method}}}
& \multicolumn{4}{c}{\hspace*{5pt}\textbf{Preservation}}
&
& \makebox[0pt][c]{\textbf{Semantic}}
&
& \makebox[0pt][c]{\textbf{Overall}} \\

\cmidrule(l{4pt}r{4pt}){2-5}
\cmidrule(l{10pt}r{10.5pt}){6-8}
\cmidrule(l{2pt}r{2pt}){9-9}

&
{SSIM$\uparrow$}
& {PSNR$\uparrow$}
& {DISTS$\downarrow$}
& {LPIPS$\downarrow$}
&
& {\hspace*{1.5pt}CLIP$\uparrow$}
&
& {\ Judge$\uparrow$} \\
\midrule

\textbf{InstructPix2Pix}~\cite{pix2pix}
& 0.67
& 16.58
& 0.20
& 0.46
&
& \textbf{26.68}
&
& 2.69 \\

\hspace{1.2em}\textbf{+SR-Edit (Ours)}
& \cellcolor{oursgray}\textbf{0.68}
& \cellcolor{oursgray}\textbf{16.94}
& \cellcolor{oursgray}\textbf{0.18}
& \cellcolor{oursgray}\textbf{0.43}
& \cellcolor{oursgray}
& \cellcolor{oursgray}25.07
& \cellcolor{oursgray}
& \cellcolor{oursgray}\textbf{2.75} \\

\midrule

\textbf{AnyEdit}~\cite{anyedit}
& 0.70
& 18.95
& 0.17
& 0.41
&
& 25.15
&
& 2.82 \\

\hspace{1.2em}\textbf{+SR-Edit (Ours)}
& \cellcolor{oursgray}\textbf{0.85}
& \cellcolor{oursgray}\textbf{19.19}
& \cellcolor{oursgray}\textbf{0.13}
& \cellcolor{oursgray}\textbf{0.35}
& \cellcolor{oursgray}
& \cellcolor{oursgray}\textbf{25.21}
& \cellcolor{oursgray}
& \cellcolor{oursgray}\textbf{2.99} \\

\midrule

\textbf{Qwen-Image-Edit} 2511~\cite{qwen-image}
& 0.61
& 14.92
& 0.23
& 0.46
&
& \underline{26.16}
&
& 3.84 \\

\hspace{1.2em}\textbf{+Follow-Your-Shape}~\cite{follow-your-shape}
& 0.61
& 14.96
& 0.23
& 0.47
&
& 26.03
&
& 3.59 \\

\hspace{1.2em}\textbf{+SpotEdit}~\cite{spot-edit}
& \textbf{0.70}
& \underline{16.55}
& \underline{0.20}
& \underline{0.32}
&
& 26.11
&
& \underline{3.91} \\

\hspace{1.2em}\textbf{+SR-Edit (Ours)}
& \cellcolor{oursgray}\underline{0.67}
& \cellcolor{oursgray}\textbf{17.40}
& \cellcolor{oursgray}\textbf{0.18}
& \cellcolor{oursgray}\textbf{0.31}
& \cellcolor{oursgray}
& \cellcolor{oursgray}\textbf{26.80}
& \cellcolor{oursgray}
& \cellcolor{oursgray}\textbf{3.94} \\

\midrule

\textbf{Step1X-Edit} v1p2~\cite{step1x}
& 0.68
& 15.96
& 0.20
& 0.39
&
& 25.89
&
& 4.00 \\

\hspace{1.2em}\textbf{+Follow-Your-Shape}
& 0.67
& 15.93
& 0.20
& 0.39
&
& 25.84
&
& \underline{4.03} \\

\hspace{1.2em}\textbf{+SpotEdit}
& \underline{0.75}
& \textbf{16.77}
& \underline{0.17}
& \underline{0.31}
&
& \underline{25.91}
&
& \textbf{4.08} \\

\hspace{1.2em}\textbf{+SR-Edit (Ours)}
& \cellcolor{oursgray}\textbf{0.77}
& \cellcolor{oursgray}\underline{16.48}
& \cellcolor{oursgray}\textbf{0.14}
& \cellcolor{oursgray}\textbf{0.27}
& \cellcolor{oursgray}
& \cellcolor{oursgray}\textbf{26.09}
& \cellcolor{oursgray}
& \cellcolor{oursgray}4.01 \\

\cmidrule[\heavyrulewidth](l{-2pt}r{-2pt}){1-9}
\end{tabular}

\caption{\textbf{Comparison between different methods on ImgEdit-Bench.}
\textbf{Bold} indicates the best result for each metric within each backbone group,
and \underline{underlining} indicates the second best.
For the first two backbone groups, we do not mark second best results.
CLIP scores are reported after multiplication by $10^2$.}
\label{tab:results}
\end{table}

\subsection{Backbone Editors}
\label{sec:backbones}
We test our method on four open-source backbones from two editor families.
For \emph{diffusion-based} backbones, we include
\textbf{InstructPix2Pix}, a standard instruction-following diffusion editor~\cite{pix2pix}, and \textbf{AnyEdit}, a strong unified diffusion image editor that shows competitive editing performance across diverse instructions~\cite{anyedit}.
For \emph{flow-based} editors, \textbf{Qwen-Image-Edit} 2511 and \textbf{Step1X-Edit} v1p2 are recent flow-style editors that are commonly adopted as modern backbones for general instruction-based editing~\cite{qwen-image,step1x}.

\subsection{Baselines}
\label{baselines}
We compare with two representative training-free region-aware methods that automatically infer editable regions and enforce preservation in non-edit areas during inference.
\textbf{Follow-Your-Shape} uses a trajectory divergence map to localize editable regions and applies scheduled KV~\cite{attention} injection to preserve non-target content during editing~\cite{follow-your-shape}. 
\textbf{SpotEdit} identifies editable regions via reconstruction-based stability estimation and preserves context using KV caching with interpolation from the source image~\cite{spot-edit}.

\begin{table}[b!]
\centering
\renewcommand{\arraystretch}{1.1}
\setlength{\abovecaptionskip}{8pt}
\setlength{\tabcolsep}{3.8pt}

\begin{tabular}{
    l
    *{4}{w{c}{0.85cm}}
    @{}p{12pt}@{}
    *{2}{w{c}{1.1cm}}
}
\toprule

\multirow{2}{*}{\raisebox{-0.6ex}{\textbf{Method}}}
& \multicolumn{4}{c}{\hspace*{8pt}\textbf{Preservation}}
& & \multicolumn{2}{c}{\hspace*{1.5pt}\textbf{Semantic}} \\

\cmidrule(l{5pt}r{0pt}){2-5}
\cmidrule(l{17pt}r{2pt}){6-8}

& {SSIM$\uparrow$}
& {MSE$\downarrow$}
& {SD$\downarrow$}
& {LPIPS$\downarrow$}
& &
{\ \ CLIP$_W$$\uparrow$}
& {\ \ CLIP$_E$$\uparrow$} \\

\midrule

\textbf{InstructPix2Pix}~\cite{pix2pix}
& 0.76
& 2.26
& 5.93
& 15.57
& &
29.30
& 25.89 \\

\textbf{\hspace{1.2em}+SR-Edit (Ours)}
& \cellcolor{oursgray}\textbf{0.86}
& \cellcolor{oursgray}\textbf{1.81}
& \cellcolor{oursgray}\textbf{4.30}
& \cellcolor{oursgray}\textbf{10.45}
& \multicolumn{1}{@{}>{\columncolor{oursgray}[0pt][0pt]}p{15pt}@{}}{}
& \cellcolor{oursgray}\textbf{29.79}
& \cellcolor{oursgray}\textbf{26.02} \\

\midrule

\textbf{Qwen-Image-Edit 2511}~\cite{qwen-image}
& 0.87
& 0.88
& 4.91
& 8.22
& &
31.45
& \textbf{26.99} \\

\textbf{\hspace{1.2em}+SR-Edit (Ours)}
& \cellcolor{oursgray}\textbf{0.94}
& \cellcolor{oursgray}\textbf{0.72}
& \cellcolor{oursgray}\textbf{4.50}
& \cellcolor{oursgray}\textbf{4.69}
& \multicolumn{1}{@{}>{\columncolor{oursgray}[0pt][0pt]}p{15pt}@{}}{}
& \cellcolor{oursgray}\textbf{31.46}
& \cellcolor{oursgray}26.90 \\

\bottomrule
\end{tabular}

\caption{\textbf{Additional quantitative results on PIE-Bench.}
SSIM, LPIPS, MSE are computed over human-annotated non-edit regions. SD measures overall structural preservation, and is computed on the  entire image. CLIP$_W$ and CLIP$_E$ measure CLIP similarity of target text description with the whole image and the edited region, respectively.
LPIPS, MSE, SD, and CLIP scores are reported after multiplication by $10^2$.}

\label{tab:pie-results}
\end{table}

\subsection{Metrics}
\label{sec:metrics}
We report a CLIP-based text--image similarity score to reflect whether the edited result is semantically consistent with the instruction~\cite{clip}. 
We report PSNR, SSIM, DISTS, and LPIPS between the edited image and the source image to measure preservation and fidelity, where PSNR and SSIM emphasize structure similarity, while DISTS and LPIPS capture perceptual distance~\cite{ssim,dists,lpips}.
We also include the ImgEdit judge-model score as an automatic evaluator that is designed to align with human preference for instruction-based editing~\cite{imgedit}.

For PIE-Bench, we report SSIM, LPIPS, and MSE over the human-annotated non-edit regions for region-specific preservation evaluation, together with Structure Distance (SD)~\cite{pnp} for structural preservation. We further report CLIP-Whole and CLIP-Edited to assess semantic consistency at the whole-image and edit-region levels, respectively.

\subsection{Inference Configuration}
Unless otherwise specified, we use the \emph{default} settings provided by each baseline model and method. 
For {SpotEdit}~\cite{spot-edit} and {Follow-Your-Shape}~\cite{follow-your-shape}, when the inference step count differs from the backbone editor's setting, we \emph{linearly rescale} the editing schedule to match the new number of steps.
For {SR-Edit}, we instantiate two iterations and set few-step inference stride to 3. The post-process parameters are set to $s_\text{open}=3,s_\text{close}=5,A_{\min}=1000$. We set $\sigma_\text{obs} = 3\times 10^{-2}$, and employ a linearly scheduled $h$-transform guidance strength. We enable self-refinement starting at $30\%$ of the model's inference steps.

\section{Results}

\subsection{Quantitative Results}

\begin{figure}[t!]
    \centering
    \includegraphics[width=\linewidth]{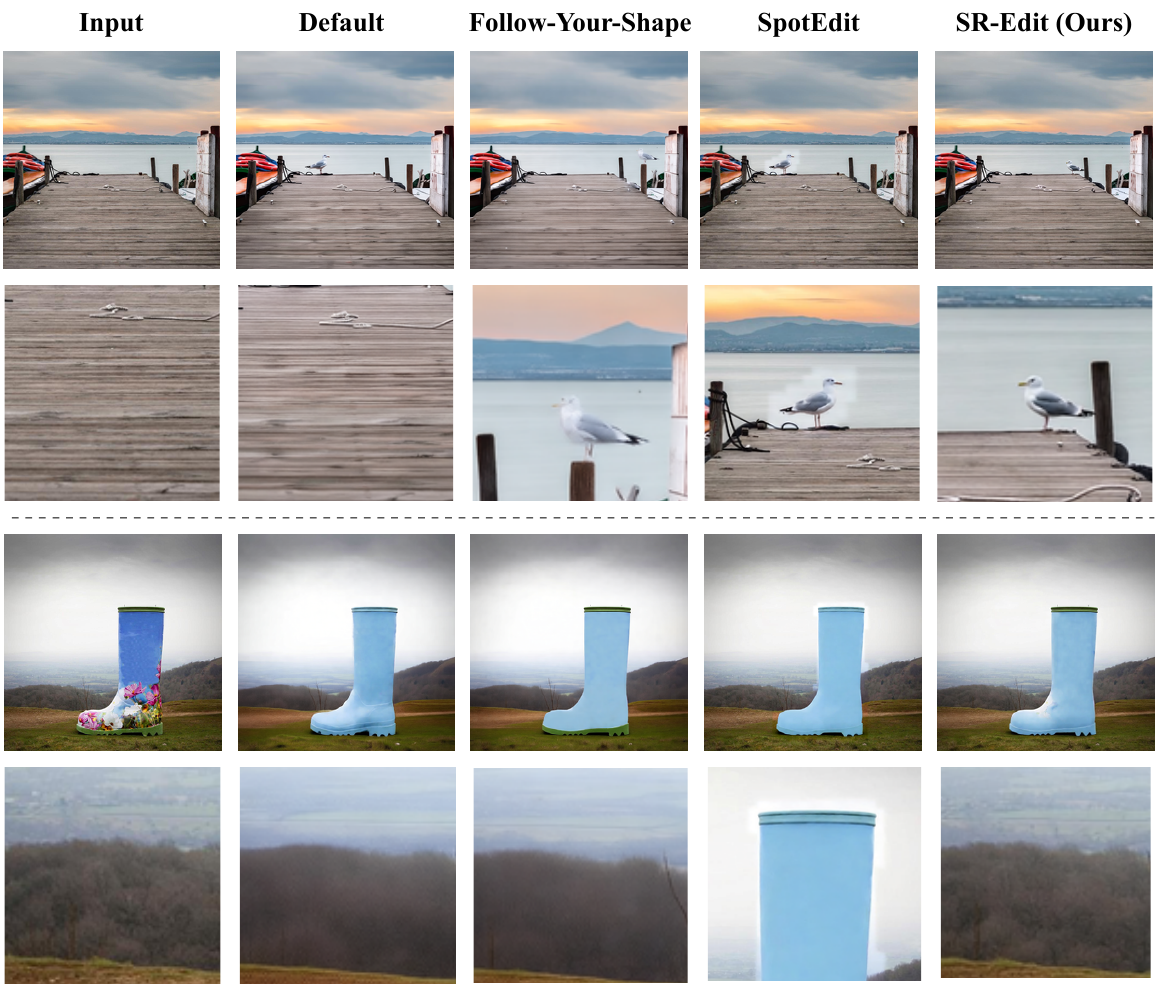}
    \caption{\textbf{Qualitative comparison}. Many baselines (\textit{e.g.} the default setting) show weaker preservation to the source image, and Follow-Your-Shape can yield insufficiently realistic edits with residual blur. SpotEdit often introduces boundary discontinuities (edge jumps). In contrast, our method is more faithful to the input and produces realistic, high-quality edits. 
    The instructions of the two cases are \emph{Add a seagull perched on the edge of the wooden pier} and \emph{Change the object color to a soft blue.} 
    All editing methods use Qwen-Image-Edit 2511~\cite{qwen-image} as backbone.}
    \label{fig:qual-comp}
\end{figure}

\begin{table}[t!]
\centering
\renewcommand{\arraystretch}{1.1}
\setlength{\abovecaptionskip}{8pt}
\setlength{\tabcolsep}{5.2pt}

\caption{\textbf{Ablation results on ImgEdit-Bench.} All methods use Qwen-Image-Edit 2511~\cite{qwen-image} as the backbone. \textbf{Reconstruction} uses the model's one-step reconstruction output for region estimation, while \textbf{Single Iter} uses only one iteration, meaning the edit region is identified once and then kept fixed. \textbf{Constant} replaces the default linearly increasing preservation guidance with a constant strength, and \textbf{Linear Down} reverses the schedule by linearly decreasing the guidance strength over the correction process.} 
\label{tab:ablation-inference}

\begin{tabular}{l*{4}{w{c}{1.08cm}}@{}p{13pt}@{}c@{}p{17pt}@{}c}
\cmidrule[\heavyrulewidth](l{-2pt}r{-1.5pt}){1-9}

\multirow{2}{*}{\raisebox{-0.5ex}{\textbf{Method}}}
& \multicolumn{4}{c}{\textbf{Preservation}}
&
& \makebox[0pt][c]{\textbf{Semantic}}
&
& \makebox[0pt][c]{\textbf{Overall}} \\

\cmidrule(l{8pt}r{2pt}){2-5}
\cmidrule(l{12pt}r{13pt}){6-8}
\cmidrule(l{2pt}r{6pt}){9-9}

&
{SSIM$\uparrow$}
& {PSNR$\uparrow$}
& {DISTS$\downarrow$}
& {LPIPS$\downarrow$}
&
& {\hspace*{2pt}CLIP$\uparrow$}
&
& {\ Judge$\uparrow$} \\

\midrule

\textbf{SR-Edit}
& \cellcolor{oursgray}\textbf{0.67}
& \cellcolor{oursgray}\textbf{17.40}
& \cellcolor{oursgray}\textbf{0.18}
& \cellcolor{oursgray}\textbf{0.31}
& \multicolumn{1}{@{}>{\columncolor{oursgray}[0pt][0pt]}p{15pt}@{}}{}
& \cellcolor{oursgray}26.80
& \multicolumn{1}{@{}>{\columncolor{oursgray}[0pt][0pt]}p{15pt}@{}}{}
& \cellcolor{oursgray}\textbf{3.94} \\

\textbf{Reconstruction}
& 0.64
& 16.90
& 0.21
& 0.35
&
& 26.95
&
& 3.88 \\

\textbf{Single Iter}
& 0.60
& 16.10
& 0.24
& 0.41
&
& 25.90
&
& 3.60 \\

\textbf{Constant}
& 0.62
& 16.56
& 0.22
& 0.38
&
& 26.97
&
& \textbf{3.94} \\

\textbf{Linear Down}
& 0.61
& 16.37
& 0.24
& 0.40
&
& \textbf{27.03}
&
& 3.87 \\

\cmidrule[\heavyrulewidth](l{-2pt}r{-2pt}){1-9}
\end{tabular}
\vspace{-0.6cm}
\end{table}

\begin{figure*}[b!]
  \vspace{-0.1cm}
  \centering
  \begin{minipage}[t]{0.50\textwidth}
\ \\
    \centering
    \includegraphics[width=0.98\linewidth]{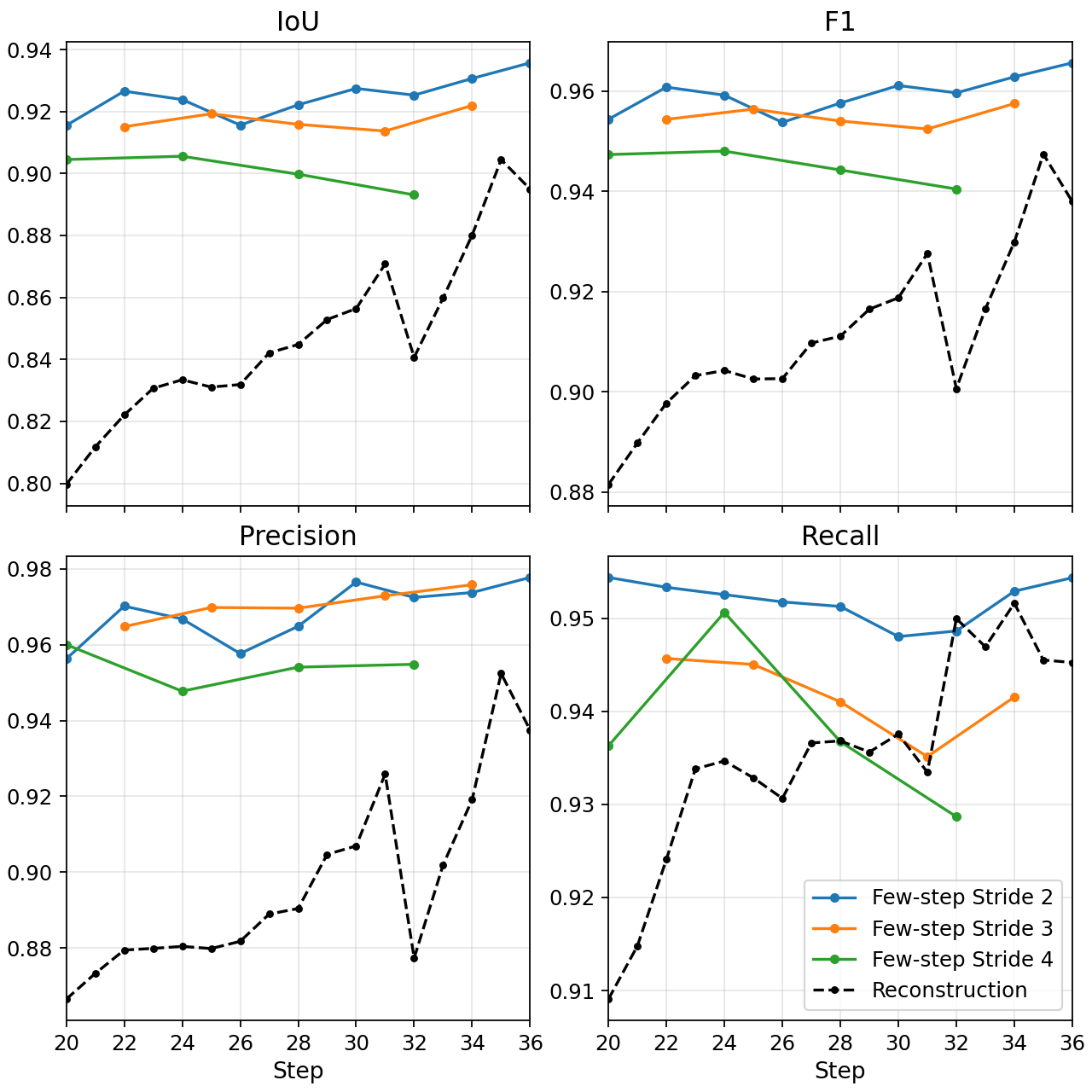}
    \caption{\textbf{Comparison between few-step inference and reconstruction}. Few-step inference is better at almost all steps.}
    \label{fig:ablation-figure}
  \end{minipage}\hfill
  \begin{minipage}[t]{0.47\textwidth}
\ \\
    \centering
    {\renewcommand{\arraystretch}{1.09}
    {\setlength{\tabcolsep}{4.2pt}
    \begin{tabular}{lcccc}
      \toprule
      Setting & P & R & IoU & F1 \\
      \midrule
      w/o Otsu  & 0.10 & 1.00 & 0.10 & 0.16 \\
      w/o Open  & 0.69 & 1.00 & 0.69 & 0.76 \\
      w/o Close & 1.00 & 0.91 & 0.91 & 0.95 \\
      w/o CC    & 0.76 & 1.00 & 0.76 & 0.84 \\
      \bottomrule
    \end{tabular}}}
    \captionof{table}{\textbf{Ablation results on post-processing pipeline of region identification}. The results indicate that each component contributes in a distinct and complementary manner. For example, removing the \emph{opening} module may increase recall, as fewer constraints allow more positive predictions, but this comes at the expense of precision and overall balance. These variations confirm that each module plays a specific role in improving the overall accuracy and robustness.} 
    \label{tab:ablations}
  \end{minipage}
\end{figure*}

As shown in Tab.~\ref{tab:results}, SR-Edit consistently improves source preservation across different backbone models while largely maintaining semantic alignment. For example, SSIM increases from $0.70$ to $0.85$ on AnyEdit and from $0.68$ to $0.77$ on Step1X-Edit v1p2, while LPIPS is reduced from $0.41$ to $0.35$ and from $0.39$ to $0.27$, respectively. On Qwen-Image-Edit 2511, SR-Edit also achieves the best PSNR, DISTS, and LPIPS while improving CLIP from $26.16$ to $26.80$, showing that stronger preservation does not generally weaken the intended edit.

PIE-Bench provides more direct region-level evidence for this behavior. SR-Edit improves non-edit SSIM from $0.76$ to $0.86$ on InstructPix2Pix and from $0.87$ to $0.94$ on Qwen-Image-Edit, with clear reductions in LPIPS, MSE, and structural distance. Meanwhile, semantic scores are either improved or largely preserved, indicating that SR-Edit mainly suppresses unintended changes outside the target region rather than simply producing more conservative edits.

\subsection{Qualitative Results}
Fig.~\ref{fig:qual-comp} shows that our method achieves higher fidelity to the input and more realistic, higher-quality edits than competing methods.

\section{Ablation Study}
\label{sec:ablation}

\paragraph{Few-step inference versus reconstruction for region estimation.}
We compare few-step inference under different strides with reconstruction in terms of region-identification quality (Fig.~\ref{fig:ablation-figure}).
Overall, few-step inference consistently outperforms reconstruction in both IoU and F1.
Moreover, reconstruction shows noticeable instability (e.g., a sharp drop around step 32 in both metrics), whereas the few-step curves remain smooth.
Among the few-step settings, stride-2 yields the best performance; stride-3 is slightly lower but close; and stride-4 tends to underperform, as recall degrades more substantially with increasing steps.
In practical self-refined inference, using regions inferred via reconstruction results in less faithful edits (Tab.~\ref{tab:ablation-inference}).

\begin{table}[t!]

\caption{\textbf{Runtime and memory analysis}. ``Time'' denotes the average inference time (s) over 100 randomly sampled cases from our ImgEdit-Bench evaluation dataset, and ``VRAM'' denotes the peak GPU memory usage (GB) during inference. All analyses are performed using a single NVIDIA RTX PRO 6000 GPU. ``Vanilla'' refers to the model's original inference. For each model, the best and second-best results among the applicable editing methods are highlighted in \textbf{bold} and \underline{underlined}, respectively. Vanilla results are provided as baselines and excluded from the ranking.}
\label{tab:runtime_vram}

\centering
\renewcommand{\arraystretch}{1.1}
\setlength{\abovecaptionskip}{10pt}
\setlength{\tabcolsep}{0.9pt}
\begin{tabular}{l@{\hspace{6pt}}cccccccc}
\toprule
\multirow{2}{*}{\textbf{Model}}
& \multicolumn{2}{c}{\makebox[0pt][c]{\textbf{Vanilla}}}
& \multicolumn{2}{c}{\makebox[5pt][c]{\hspace*{0pt}\textbf{Follow-Your-Shape}}}
& \multicolumn{2}{c}{\makebox[0pt][c]{\hspace*{0pt}\textbf{SpotEdit}}}
& \multicolumn{2}{c}{\makebox[0pt][c]{\textbf{SR-Edit}}} \\
\cmidrule(lr){2-3}
\cmidrule(lr){4-5}
\cmidrule(lr){6-7}
\cmidrule(lr){8-9}
& Time & VRAM &\ \ \ Time\ \ &\ \ VRAM\ \ \ \ & Time & VRAM & Time & VRAM \\
\midrule
InstructPix2Pix &  2.49 &  3.55 & \multicolumn{4}{c}{\multirow{2}{*}[0em]{\textit{Not Applicable}}} & \cellcolor{oursgray}\textbf{3.41} & \cellcolor{oursgray}\textbf{3.55} \\
AnyEdit &  9.84 & 12.93 & \multicolumn{4}{c}{} & \cellcolor{oursgray}\textbf{13.49} & \cellcolor{oursgray}\textbf{12.93} \\
Qwen-Image-Edit 2511    & 53.69 & 59.94 & 104.37 & \underline{62.89} & \textbf{24.60} & 66.35 & \cellcolor{oursgray}\underline{75.31} & \cellcolor{oursgray}\textbf{59.94} \\
Step1X-Edit v1p2  & 70.81 & 42.41 & 139.01 & 44.88 & \textbf{26.76} & \underline{44.77} & \cellcolor{oursgray}\underline{97.45} & \cellcolor{oursgray}\textbf{42.41} \\
\bottomrule
\end{tabular}
\end{table}

\paragraph{Iterative refinement.}
Next, we ablate the number of iterative self-refinement blocks.
Using two SR-Edit blocks improves region alignment, as the mask and preservation guidance are re-estimated from progressively refined predictions.
With a single block, the mask is fixed early and is typically over-inclusive and inaccurate, which weakens non-edit preservation and may also reduce instruction adherence by over-constraining the model (Tab.~\ref{tab:ablation-inference}).
These results highlight the importance of iterative mask re-estimation for simultaneously maintaining preservation and edit expressiveness.

\begin{wrapfigure}[27]{r}{0.46\textwidth}
    \centering
    \includegraphics[width=\linewidth]{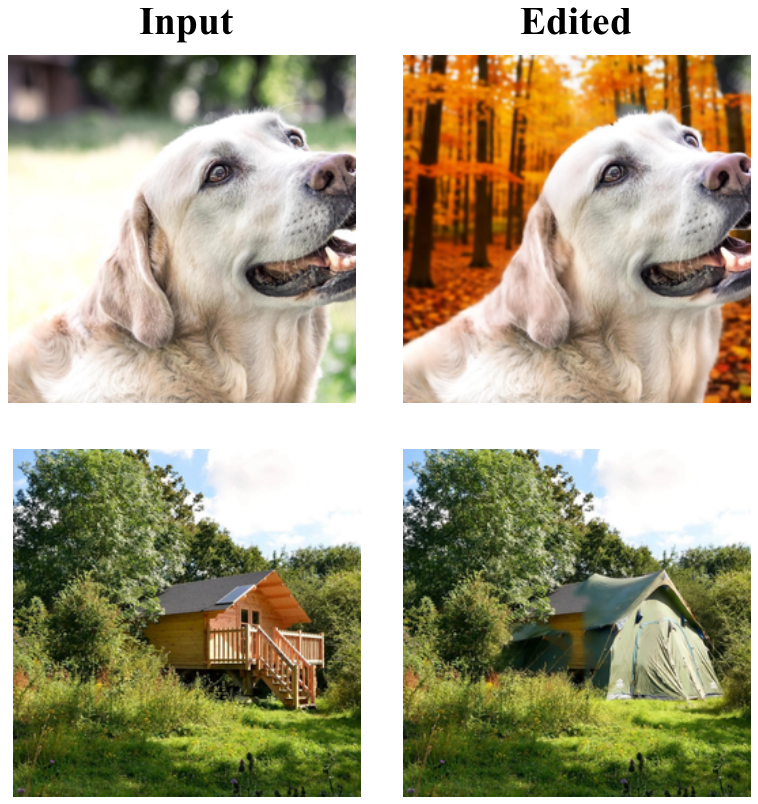}
    \caption{SR-Edit sometimes introduces unintended blending between the preserved content and the target edit. The first example shows a dark artifact in the upper-right corner, while the second produces a hybrid cabin–tent structure due to incomplete replacement. The edit instructions for the two cases are \textit{Change the blurred environment in the background to an autumn forest with orange and
    yellow leaves on the trees} and  \textit{Replace the wooden cabin in the image with a large camping tent}.}
    \label{fig:fail}
\end{wrapfigure}

\paragraph{Region post-processing pipeline.}
We ablate each post-processing component against the full pipeline output to examine its role in region identification.
Tab.~\ref{tab:ablations} indicates that the four steps are complementary: removing Otsu produces an almost non-informative mask with extremely low precision; removing opening increases false positives and lowers precision; removing closing harms spatial coherence and reduces overlap, and removing connected-component filtering retains noisy fragments and degrades overall balance.
Together, these results demonstrate the distinct yet complementary roles of the individual components.

\paragraph{Guidance schedule.}
As shown in Tab.~\ref{tab:ablation-inference}, the linearly increasing schedule consistently improves preservation over \textbf{Constant} and \textbf{Linear Down}, with minor variation in semantic and overall quality. One key factor underlying this trend is the decreasing uncertainty of $p(\bm x_0\mid\bm x_t,\mathcal C)$: at noisy steps, its larger covariance yields a smaller inverse covariance and thus weaker effective preservation, while the increasing precision later in sampling strengthens the correction. The increasing schedule we adopt captures this tendency and is therefore better aligned with the actual theoretical correction.

\vspace{-8pt}
\section{Computational Efficiency Analysis}
\vspace{-2pt}

As shown in Tab.~\ref{tab:runtime_vram}, SR-Edit introduces negligible additional memory overhead, with peak VRAM usage remaining essentially unchanged from the vanilla setting. While it introduces a moderate increase in inference time, this additional cost is generally acceptable considering the substantial improvement in source fidelity. More importantly, SR-Edit can be used to construct large-scale, high-fidelity editing pairs, allowing its source-preserving capability to be distilled into the base model. This provides a practical path toward retaining the benefits of SR-Edit while eliminating the additional sampling overhead at inference time.

\vspace{-8pt}
\section{Scope and Limitations}
\vspace{-2pt}

We currently focus on localized editing, which is the primary target setting of SR-Edit. Our main observed failures arise when preservation interferes with the intended edit semantics: (1) over-estimating non-edit region, leading to abnormal mix between guidance and edit modification (Fig.~\ref{fig:fail}). (2) preservation may still be slightly weaker than hard replacement, as our soft correction cannot fully reset drifted non-edit latents, allowing residual interference to propagate into the edited region and leading to semantic failures. As stronger preservation is the central goal of this work, we will continue refining the design to improve preservation while maintaining the integrity of the intended edit semantics.

\section{Conclusion}

In this paper, we present SR-Edit, a region-aware image editing method that performs faithful edits without external masks. It uses iterative self-refinement to infer and sharpen edit regions from pixel-space predictions, and applies a Doob's $h$-transform–based masked residual correction to improve preservation for diffusion and flow-based backbones. Experiments across multiple editors and metrics show better non-edit preservation and visual quality than representative baselines while maintaining instruction consistency. Ablation studies further verify the benefits of few-step probing for robust region estimation and the necessity of iterative mask re-estimation for balancing preservation and edit expressiveness. Overall, SR-Edit offers a simple, training-free framework that can be applied to modern backbones to reduce unintended drift during inference.

\section{Acknowledgment}
This work is supported by the National Natural Science Foundation of China (62550004, U24A20342, U25B6003, 92570001).

\bibliographystyle{splncs04}
\bibliography{main}

\newpage

\begin{center}
    \Large\textbf{
        SR-Edit: Region-Aware Image Editing via Self-Refinement\\[0.4em]
        -- Supplementary Material
    }
\end{center}

\setcounter{section}{0}

\section{Pseudo Code of Post-Processing and Self-Refinement}

We provide the pseudo code of the proposed SR-Edit framework below. 
It consists of a pixel-space post-processing pipeline for region identification and an iterative self-refinement 
inference procedure. 

\begin{algorithm}[h!]
\caption{Edit Region Identification via Pixel-Space Post-Processing}
\label{alg:region_identification}
\begin{algorithmic}[1]
\Require Input image $\mathbf{I}_{\mathrm{in}} \in \mathbb{R}^{H \times W \times C}$;
         provisional edited image $\mathbf{I}_{\mathrm{ref}} \in \mathbb{R}^{H \times W \times C}$;
         opening size $s_{\mathrm{open}}$;
         closing size $s_{\mathrm{close}}$;
         minimum area $A_{\min}$

\ForAll{$p \in \Omega$}
    \State Compute the difference map
$
    d(p) \gets \frac{1}{C}\sum_{c=1}^{C}
    \left|\mathbf{I}_{\mathrm{in}}(p,c)-\mathbf{I}_{\mathrm{ref}}(p,c)\right|
$
\EndFor

\State Compute the Otsu threshold~\cite{otsu}
$
t^\star \gets \arg\max_t \ \omega_0(t)\omega_1(t)\bigl(\mu_0(t)-\mu_1(t)\bigr)^2
$

\ForAll{$p \in \Omega$}
    \State $\textbf{M}_0(p) \gets \mathbb{I}[\, d(p) \ge t^\star \,]$
\EndFor

\State Construct circular structuring elements~\cite{morphological,morphological2} $B_{\mathrm{open}}$ and $B_{\mathrm{close}}$
\State $\textbf{M}_1 \gets \mathrm{Opening}(\textbf{M}_0, B_{\mathrm{open}})$
\State $\textbf{M}_1 \gets \mathrm{Closing}(\textbf{M}_1, B_{\mathrm{close}})$

\State Compute connected components~\cite{digital-picture-processing} $\{\mathcal{C}_k\}_k$ of $\textbf{M}_1$
\State Initialize $\textbf{M}(p)\gets 0$ for all $p\in\Omega$
\ForAll{connected components $\mathcal{C}_k$}
    \If{$\mathrm{Area}(\mathcal{C}_k) \ge A_{\min}$}
        \ForAll{$p \in \mathcal{C}_k$}
            \State $\textbf{M}(p) \gets 1$
        \EndFor
    \EndIf
\EndFor

\State \Return $\textbf{M}$
\end{algorithmic}
\end{algorithm}

\begin{algorithm}[h!]
\caption{Unified SR-Edit Inference for Diffusion and Flow Backbones}
\label{alg:sredit}
\begin{algorithmic}[1]
\Require Input image $\textbf{I}_{\mathrm{in}}$;
         editing condition $\mathcal{C}$;
         terminal time $T$;
         self-correction start time $s$;
         correction duration $n$;
         number of self-refinement blocks $N$;
         probe stride $K$

\State Encode the input image into latent $\bm{x}_{\mathrm{src}}$ and initialize $\bm{x}_T$
\State Run stabilization inference from $T$ to $s$

\For{$i=0$ to $N-1$}
    \State \textbf{I. Few-step probe for region estimation}
    \State Starting from latent state at $s-i\cdot n$, run few-step inference with stride $K$ and decode the latent prediction to provisional image $\textbf{I}_{\mathrm{ref}}$
    \State Compute pixel-space mask $\textbf{M} = \textsc{RegionIdentification}(\textbf{I}_{\mathrm{in}}, \textbf{I}_{\mathrm{ref}})$
    \State Downsample the mask to latent space: ${\mathbf{M}}_{\text{latent}} = \text{Downsample}({\mathbf{M}})$

    \State \textbf{II. Self-correction update}
    \State Define the non-edit mask as $\bar{\mathbf{M}}_{\text{latent}} = 1-\mathbf{M}_{\text{latent}}$
    \State Compute latent residual guidance~\cite{doob,sb,flow-guidance}:
    $\bm{r}_t = - \, \bar{\mathbf{M}}_{\text{latent}} \bigl(\widehat{\bm{x}}_0(\bm{x}_t) - \bm{x}_{\mathrm{src}}\bigr)$
    \State Augment score estimation~\cite{ddpm,sde} or vector field~\cite{flow-matching,reci-flow} with $\bm{r}_t$ and schedule
    \State Perform self-correction sampling over $[\,s-(i+1)\cdot n,\; s-i\cdot n\,]$
\EndFor

\State Decode final latent state to $\textbf{I}_{\mathrm{out}}$  and return $\textbf{I}_{\mathrm{out}}$
\end{algorithmic}
\end{algorithm}

Algorithm~\ref{alg:region_identification} summarizes the former, 
including difference-map construction, thresholding~\cite{otsu}, morphological 
refinement~\cite{morphological,morphological2}, and 
connected-component filtering~\cite{digital-picture-processing}, while Algorithm~\ref{alg:sredit} presents the latter 
for both diffusion and flow backbones. Together, they provide a concise summary 
of the overall pipeline and practical guidance for implementation.

\section{Additional Details}
\paragraph{Quantitative analysis.}
For the analysis in Fig.~\ref{fig:change_concentration}, we use the human-annotated masks from MagicBrush test split~\cite{magicbrush} and further manually select only samples whose masks accurately match the edited region, excluding cases where the mask does not define a realizable target for a mask-free editing model (e.g., task \textit{add}), yielding 198 samples in total. Raw results are reported in Tab.~\ref{tab:noise-diff} and~\ref{tab:energy-connect}.

\begin{center}
 \begin{minipage}{0.36\textwidth}
  \centering
  \captionof{table}{Diff on level K~\cite{mad,alter-mad}}
  \vspace{0.3em}
  \label{tab:noise-diff}
  \begin{tabular*}{\linewidth}{@{\extracolsep{\fill}}ccc@{}}
  \toprule
  K & Diff (Edit) & Diff (Non) \\
  \midrule
  0.00 & 0.15 & 0.06 \\
  0.50 & 0.13 & 0.05 \\
  1.00 & 0.12 & 0.04 \\
  2.00 & 0.10 & 0.02 \\
  3.00 & 0.08 & 0.01 \\
  \bottomrule
  \end{tabular*}
 \end{minipage}
 \hfill
 \begin{minipage}{0.615\textwidth}
  \centering
  \captionof{table}{Connectivity at different energy coverages}
  \vspace{0.3em}
  \label{tab:energy-connect}
  \begin{tabular*}{\linewidth}{@{\extracolsep{\fill}}ccccc@{}}
  \toprule
  Energy & CC(Edit) & CC(Non) & LCC(Edit) & LCC(Non) \\
  \midrule
  0.50 & 100.45 & 1472.12 & 0.64 & 0.29 \\
  0.60 & 104.85 & 1741.81 & 0.67 & 0.31 \\
  0.70 & 99.42 & 1912.12 & 0.70 & 0.36 \\
  0.80 & 87.44 & 1934.48 & 0.76 & 0.47 \\
  0.90 & 77.43 & 1664.87 & 0.85 & 0.69 \\
  \bottomrule
  \end{tabular*}
 \end{minipage}
\end{center}

\section{Sensitivity and Robustness}
\begin{figure}[htbp]
    \centering
    \includegraphics[width=\linewidth]{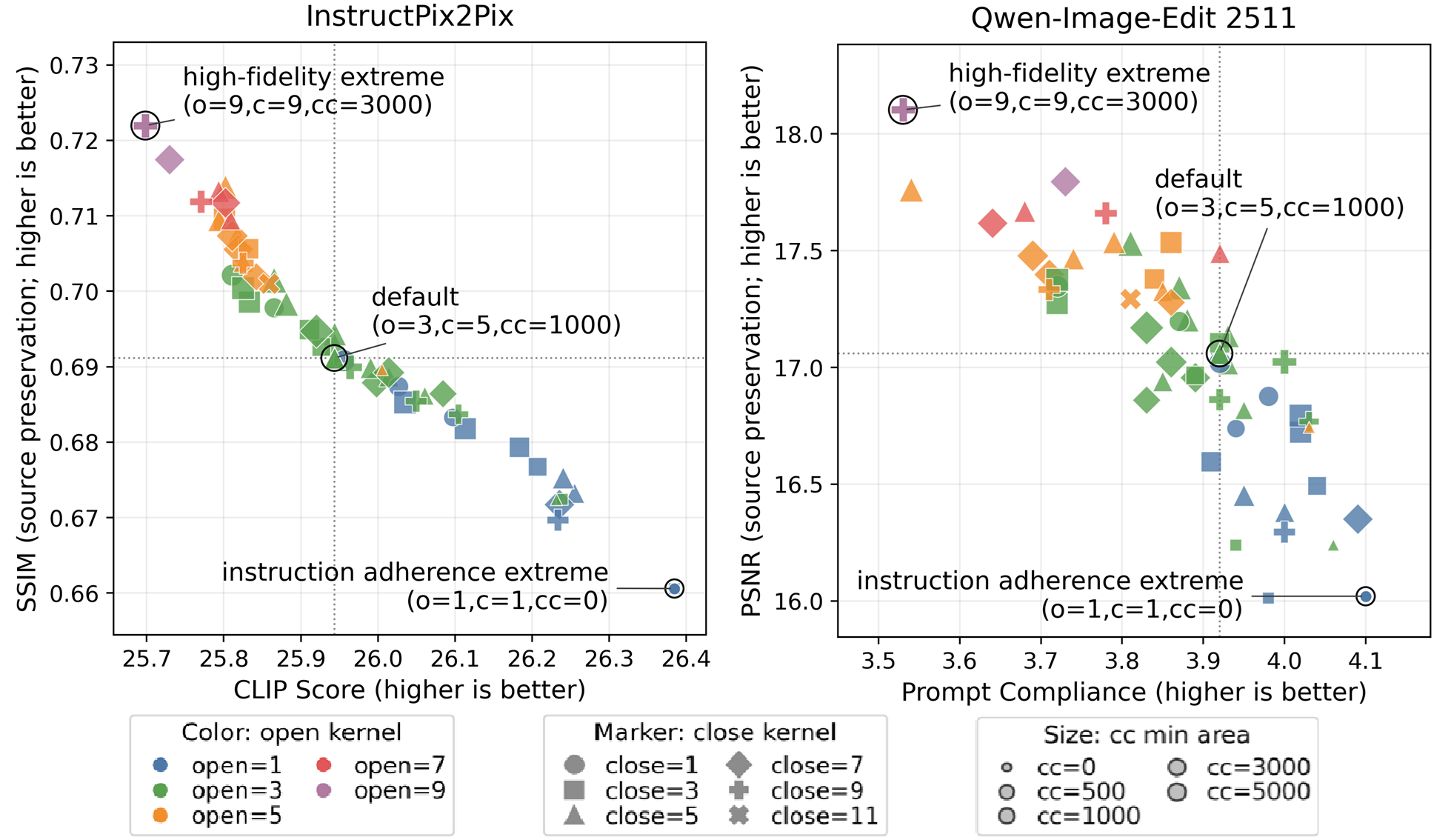}
    \caption{\textbf{Sensitivity of mask parameters on 54 configs over 100 cases randomly chosen from our ImgEdit-Bench evaluation dataset.} ``Prompt compliance'' denotes instruction-adherence score from Judge~\cite{imgedit} model.}
    \label{fig:sensitivity}
\end{figure}

\paragraph{Mask building.} In Fig.~\ref{fig:sensitivity}, across different backbones and metrics, the results show a clear, reasonable negative relation between source preservation and instruction adherence, while nearby configurations vary \textbf{smoothly}, indicating the strong \textbf{stability} of SR-Edit.

\vspace{-0.95em}

\paragraph{Resolution and approximation.}
In Tab.~\ref{tab:resolution_schedule}, SR-Edit robustly improves fidelity across \textbf{resolutions}, and fixed mask parameters remain effective. Since exact Doob guidance requires unavailable true ${\bm x}_0$ (data) posterior, we compare plug-in $\widehat{\bm x}_0$ with averaged rollout estimation, which is theoretically more accurate but brings no gain (also costly), suggesting \textbf{one-step estimation is sufficient}.

\begin{table}[h!]
\centering
\setlength{\tabcolsep}{3pt}
\begin{tabular}{lcccccccccc}
\toprule
\multirow{2}{*}[-0.1em]{\textbf{Metric}}
& \multicolumn{3}{c}{\textbf{Resolution} 0.5\(\times\)}
& \multicolumn{2}{c}{1.0\(\times\)}
& \multicolumn{3}{c}{1.25\(\times\)}
& \multicolumn{2}{c}{\textbf{Approx.}}\\
\cmidrule(lr){2-4}
\cmidrule(lr){5-6}
\cmidrule(lr){7-9}
\cmidrule(lr){10-11}
& Vanilla & Uns. & Scale
& Vanilla & SR-Edit
& Vanilla & Uns. & Scale
& Roll-3 & Roll-5 \\
\midrule
PSNR
& 9.56 & 10.70 & 10.01
& 14.59 & 17.06
& 12.06 & 13.70 & 14.14
& 16.80 & 16.84 \\
SSIM
& 0.25 & 0.40 & 0.32
& 0.54 & 0.70
& 0.39 & 0.62 & 0.65
& 0.61 & 0.58 \\
CLIP
& 27.27 & 27.33 & 27.42
& 26.83 & 26.99
& 27.12 & 27.27 & 27.32
& 27.39 & 26.46 \\
Judge
& 3.12 & 3.10 & 3.19
& 3.86 & 3.93
& 3.72 & 3.77 & 3.69
& 3.81 & 3.87 \\
\bottomrule
\end{tabular}

\vspace{0.6em}
\caption{\textbf{Robustness analysis on Qwen-Image-Edit 2511}~\cite{qwen-image} \textbf{on the same dataset as in the previous analysis.} ``Vanilla'' denotes the model's original inference. ``Scale'' rescales mask parameters by resolution factor, while ``Uns.'' keeps them fixed. Roll-3 and Roll-5 replace plug-in with average of 5 full-chain estimates with stride 3 and 5. CLIP scores are reported after multiplication by $10^2$.}
\label{tab:resolution_schedule}
\end{table}

\end{document}